# Triangular Fuzzy Rescaling Distance

Eddy Soria[1[0000-0001-9999-5850]], Aida Valls[2[0000-0003-3616-7809]] and Ana Beatriz Hernández[1[0000-0002-8110-9328]]

[1] Department of Business Management, Universitat Rovira i Virgili, Catalonia, Spain
eddy.soria@urv.cat, anabeatriz.hernandez@urv.cat
[2] Department of Computer Science and Mathematics, Universitat Rovira i Virgili, Catalonia, Spain
aida.valls@urv.cat

**Abstract.** Decision-making in complex systems often involves dealing with imprecise or uncertain information, frequently represented using fuzzy sets, particularly Triangular Fuzzy Numbers (*TFNs*). A crucial aspect of many fuzzy methods is the quantification of distance between TFNs. Many distance measures assume that all values are in the same scale, requiring a preliminary normalization stage when applied to heterogeneous attributes with different scales or units. This paper proposes the Triangular Fuzzy Rescaling Distance ($d_{TR}$), a metric designed to address this challenge. The $d_{TR}$ uniquely integrates Linear Rescaling (*LRE*) directly into the distance calculation, ensuring normalization during the comparison of fuzzy numbers. We formally prove that $d_{TR}$ satisfies the properties of a metric, including non-negativity, identity, symmetry, and the triangle inequality. Furthermore, we demonstrate that $d_{TR}$ is bounded, scale-invariant, and origin-invariant. These properties, combined with a weighting vector for prioritizing dimensions, make $d_{TR}$ suitable for applications involving heterogeneous fuzzy data, such as the construction of synthetic indicators, distance-based machine learning algorithms or multicriteria-decision aiding.

**Keywords:** Triangular Fuzzy Numbers, Fuzzy Distance, Data Normalization, Linear Rescaling.

## 1 Introduction

Decision-making and data analysis in complex systems often involves dealing with uncertainty and imprecision, stemming from incomplete data, subjective judgments, or inherent variability [1, 2]. Fuzzy set theory, and in particular Triangular Fuzzy Numbers (TFNs), provide a robust mathematical framework for representing and manipulating such imprecise information [3]. For instance, TFNs can effectively model expert opinions on future market trends or quantify imprecise measurements in environmental monitoring.

Distance measures between fuzzy numbers are fundamental tools in various applications, including fuzzy clustering, pattern recognition, and decision-making [4–6].

While several distance measures exist, such as Euclidean-based approaches [6] or those relying on $\alpha$-cuts [7], many do not intrinsically handle differences in the scales or origins of the fuzzy numbers being compared. This limitation can lead to distorted results, particularly when dealing with heterogeneous data where attributes have different units or ranges, which is common in the construction of synthetic indicators and other decision-support tools [8].

To address these limitations, this paper introduces the Triangular Fuzzy Rescaling Distance ($d_{TR}$), a novel metric specifically designed for comparing $n$-tuples of TFNs. $d_{TR}$ uniquely integrates Linear Rescaling (*LRE*) directly into the distance calculation, ensuring normalization before comparison. Furthermore, $d_{TR}$ is proven to be a formal metric, possessing scale and origin invariance. This combination of features allows $d_{TR}$ to handle uncertainty while mitigating the biases introduced by differing scales and units in the data.

The remainder of this paper is organized as follows. Section 2 provides the necessary mathematical preliminaries, defining fuzzy sets, TFNs, and linear rescaling. Section 3 formally introduces the $d_{TR}$ distance and proves its key properties. Section 4 and 5 present a numerical example and a case study to illustrate the calculation of $d_{TR}$, respectively. Finally, Section 6 concludes the paper and discusses future research directions. The primary contributions of this work are the proposal of the $d_{TR}$ metric, which has potential for applications involving heterogeneous fuzzy data.

# 2 Preliminaries

## 2.1 Fuzzy Numbers

**Definition 1 (Fuzzy Set).** A fuzzy set $\Psi$ in a universe of discourse $X$ is characterized by a membership function $\mu_\Psi: X \to [0,1]$. This function assigns to each element $x \in X$ a degree of membership $\mu_\Psi(x)$ that quantifies the element's belongingness to the set $\Psi$, where $\mu_\Psi(x) = 0$ indicates no membership and $\mu_\Psi(x) = 1$ indicates full membership.

**Definition 2 (Fuzzy Number).** A fuzzy number $\tilde{N}$ is a fuzzy set on $\mathbb{R}$ characterized by a membership function $\mu_{\tilde{N}}: \mathbb{R} \to [0,1]$ that satisfies:

a) **Normality**: There exists at least one $x \in \mathbb{R}$ such that $\mu_N(x) = 1$.
b) **Convexity**: For all $x_1, x_2 \in \mathbb{R}$ and $\lambda \in [0,1]$, $\mu_{\tilde{N}}(\lambda x_1 + (1-\lambda)x_2) \geq \min\{\mu_{\tilde{N}}(x_1), \mu_{\tilde{N}}(x_2)\}$.
c) **Upper Semi-Continuity**: The membership function $\mu_{\tilde{N}}(x)$ is upper semi-continuous on $\mathbb{R}$.
d) **Bounded Support**: There exists a bounded closed interval $[c,d] \subset \mathbb{R}$ such that $\mu_{\tilde{N}}(x) = 0$ for all $x \notin [c,d]$.

**Definition 3 (Triangular Fuzzy Number).** A triangular fuzzy number (TFN) $\tilde{F}$ is a specific type of fuzzy number defined by an ordered triplet $\left(f^{(1)}, f^{(2)}, f^{(3)}\right)$, where $f^{(1)} \leq f^{(2)} \leq f^{(3)}$. Its membership function $\mu_{\tilde{F}}: \mathbb{R} \rightarrow [0,1]$ is defined as:

$$\mu(x) = \begin{cases} \dfrac{x - f^{(1)}}{f^{(2)} - f^{(1)}}, & f^{(1)} \leq x \leq f^{(2)} \\ \dfrac{-x + f^{(3)}}{f^{(3)} - f^{(2)}}, & f^{(2)} \leq x \leq f^{(3)} \\ 0, & x < f^{(1)} \; or \; x > f^{(3)} \end{cases}$$

This function increases linearly from $\mu_{\tilde{F}}\left(f^{(1)}\right) = 0$ to $\mu_{\tilde{F}}\left(f^{(2)}\right) = 1$ and then decreases linearly back to $\mu_{\tilde{F}}\left(f^{(3)}\right) = 0$.

A *crisp* number *(real* number *f)* can be represented as a TFN in a degenerate form $(f, f, f)$. The membership function $\mu_{\tilde{F}}(f)$ for this degenerate triangular fuzzy number is defined as:

$$\mu_{\tilde{F}}(f) = \begin{cases} 1, & f = f^{(1)} = f^{(2)} = f^{(3)} \\ 0, & otherwise \end{cases}$$

## 2.2 Distance Function

The *distance function* is simply known as *distance* or *metric*, but this notion is extended to many types of generalized metrics, so the definition we use in this paper is according to [1]:

**Definition 4 (metric).** Let $F$ be a non-empty set. A function d: $F \times F \rightarrow \mathbb{R}$ is called a *metric* on $F$ if, for all $f, f^*, g \in F$, there holds:

1. $d(f, f^*) \geq 0$ *(non-negativity),*
2. $d(f, f^*) = 0$ *if and only if* $f = f^*$ *(identity of indiscernibles),*
3. $d(f, f^*) = d(f^*, f)$ *(symmetry),*
4. $d(f, f^*) \leq d(f, g) + d(g, f^*)$ *(triangle inequality)*

The set $F$ equipped with a metric $d$ is called a *metric space* and denoted by *(F,d)*.

## 2.3 Linear Rescaling

To introduce this concept formally, we will adopt the definition provided by [8]:

**Definition 5 (Linear Rescaling).** Let $F = \{f \in \mathbb{R} \mid a \leq f \leq b, a \neq b\}$ be a non-empty set. Linear Rescaling is a mapping $LRE: [a, b] \rightarrow [m, M]$ defined by:

$$LRE\,(f) = M - (M - m)\left(\frac{b - f}{b - a}\right) \tag{1}$$

Rescaling performs a linear transformation on the original data. From the minimum ($a$) and maximum ($b$) value of an original attribute ($f$) in the range $[a,b]$, a normalized value ($LRE\,(f)$) is obtained in a new dimensionless range $[m, M]$.

# 3 Triangular Fuzzy Rescaling Distance

## 3.1 Definition and Properties

**Definition 6 (Triangular Fuzzy Rescaling Distance).** Let $\mathcal{F}$ be the set of n-tuples of Triangular Fuzzy Numbers, defined as $\mathcal{F} = \{\tilde{f} \in \tilde{F}^n \mid \tilde{f}_i \in [a_i, b_i], [a_i, b_i] \in IN\}$, where $\tilde{f}_i = (f_i^{(1)}, f_i^{(2)}, f_i^{(3)})$, and $IN = \{[A, B] \mid A \in \mathbb{R}, B \in \mathbb{R}, A < B\}$ is a non-empty set of degenerate TFN intervals. The *Triangular Fuzzy Rescaling Distance* $d_{TR}$ of dimension $n$ is a mapping $d_{TR}$: $\mathcal{F} \times \mathcal{F} \rightarrow [0, M]$ associated with a weighting vector $\Omega$ where $\omega_i \in [0,1]$ and $\sum_{i=1}^{n} \omega_i = 1$ defined as:

$$d_{TR}\left(\tilde{f}, \tilde{f}^*\right) = d_{TR}\left((\tilde{f}_1, \dots, \tilde{f}_n), (\tilde{f}_1^*, \dots, \tilde{f}_n^*)\right)$$

$$= \frac{M}{4}\left(\sum_{i=1}^{n} \omega_i \left(\frac{|f_i^{*(1)} - f_i^{(1)}| + 2|f_i^{*(2)} - f_i^{(2)}| + |f_i^{*(3)} - f_i^{(3)}|}{b_i - a_i}\right)^{\lambda}\right)^{\frac{1}{\lambda}} \quad (2)$$

Where $\lambda \geq 1$ and $M > 0$ are parameters.

**Proposition 1 (Metric):** Let $(\mathcal{F}, d_{TR})$ be a non-empty metric space where $\mathcal{F}$ is defined as above. Then $d_{TR}$ is a *metric* or *distance function* on $\mathcal{F}$.

*Proof.* $d_{TR}$ accomplishes the following properties:

Non- negativity: $\forall\, \tilde{f}, \tilde{f}^* \in \mathcal{F},\ d_{TR}(\tilde{f}, \tilde{f}^*) \geq 0$

The proof is straightforward and thus omitted.

Identity of indiscernibles: $\forall\, \tilde{f}, \tilde{f}^* \in \mathcal{F}, d_{TR}(\tilde{f}, \tilde{f}^*) = 0 \Leftrightarrow \tilde{f} = \tilde{f}^*$

The proof is straightforward and thus omitted.

Symmetry*:* $\forall\, \tilde{f}, \tilde{f}^* \in \mathcal{F},\ d_{TR}(\tilde{f}, \tilde{f}^*) = d_{TR}(\tilde{f}^*, \tilde{f})$

The proof is straightforward and thus omitted.

Triangle inequality: $\forall\, \tilde{f}, \tilde{g}, \tilde{f}^* \in \mathcal{F},\ d_{TR}\left(\tilde{f}, \tilde{f}^*\right) \leq d_{TR}(\tilde{f}, \tilde{g}) + d_{TR}\left(\tilde{g}, \tilde{f}^*\right)$

*Proof.* Let $\tilde{f}_i = (f_i^{(1)}, f_i^{(2)}, f_i^{(3)})$, $\tilde{f}_i^* = (f_i^{*(1)}, f_i^{*(2)}, f_i^{*(3)})$ and $\tilde{g}_i = (g_i^{(1)}, g_i^{(2)}, g_i^{(3)})$ be TFNs where $\tilde{f}_i^*, \tilde{f}_i, \tilde{g}_i \in [a_i, b_i]$. We need to prove that $d_{TR}$ satisfies the triangle inequality:

$$\frac{M}{4}\left(\sum_{i=1}^{n} \omega_i \left(\frac{|f_i^{*(1)} - f_i^{(1)}| + 2|f_i^{*(2)} - f_i^{(2)}| + |f_i^{*(3)} - f_i^{(3)}|}{b_i - a_i}\right)^{\lambda}\right)^{\frac{1}{\lambda}}$$

$$\leq \frac{M}{4}\left(\sum_{i=1}^{n} \omega_i \left(\frac{|g_i^{(1)} - f_i^{(1)}| + 2|g_i^{(2)} - f_i^{(2)}| + |g_i^{(3)} - f_i^{(3)}|}{b_i - a_i}\right)^{\lambda}\right)^{\frac{1}{\lambda}}$$

$$+ \frac{M}{4}\left(\sum_{i=1}^{n} \omega_i \left(\frac{|f_i^{*(1)} - g_i^{(1)}| + 2|f_i^{*(2)} - g_i^{(2)}| + |f_i^{*(3)} - g_i^{(3)}|}{b_i - a_i}\right)^{\lambda}\right)^{\frac{1}{\lambda}} \quad (3)$$

By the well-known scalar triangle inequality ($|a - b| \leq |a - c| + |c - b|$) we can define for all fuzzy numbers with indices *i* from 1 to *n*:

$$\left|{f_i}^{*(1)} - {f_i}^{(1)}\right| \leq \left|{g_i}^{(1)} - {f_i}^{(1)}\right| + \left|{f_i}^{*(1)} - {g_i}^{(1)}\right| \quad (4)$$

$$2\left|{f_i}^{*(2)} - {f_i}^{(2)}\right| \leq 2\left(\left|{g_i}^{(2)} - {f_i}^{(2)}\right| + \left|{f_i}^{*(2)} - {g_i}^{(2)}\right|\right) \quad (5)$$

$$\left|{f_i}^{*(3)} - {f_i}^{(3)}\right| \leq \left|{g_i}^{(3)} - {f_i}^{(3)}\right| + \left|{f_i}^{*(3)} - {g_i}^{(3)}\right| \quad (6)$$

By adding up the inequalities (4), (5) and (6) we have:

$$\begin{aligned}\left|{f_i}^{*(1)} - {f_i}^{(1)}\right| + 2\left|{f_i}^{*(2)} - {f_i}^{(2)}\right| + \left|{f_i}^{*(3)} - {f_i}^{(3)}\right| &\leq \left|{g_i}^{(1)} - {f_i}^{(1)}\right| + \left|{f_i}^{*(1)} - {g_i}^{(1)}\right| \\ &+ 2\left(\left|{g_i}^{(2)} - {f_i}^{(2)}\right| + \left|{f_i}^{*(2)} - {g_i}^{(2)}\right|\right) + \left|{g_i}^{(3)} - {f_i}^{(3)}\right| \\ &+ \left|{f_i}^{*(3)} - {g_i}^{(3)}\right| \quad (7)\end{aligned}$$

Since $b_i > a_i$ (in **Definition 6**), division by $(b_i - a_i)$ preserves the inequality. Applying the λ-th root and the power λ maintains it by convexity of $x^\lambda$ for $\lambda \geq 1$. Weighted summation with $\omega_i$, satisfying $\sum_{i=1}^{n} \omega_i = 1$ also preserves the inequality. Therefore, after applying the above transformations, we can rewrite (7) as:

$$\left(\sum_{i=1}^{n} \omega_i \left(\frac{\left|{f_i}^{*(1)} - {f_i}^{(1)}\right| + 2\left|{f_i}^{*(2)} - {f_i}^{(2)}\right| + \left|{f_i}^{*(3)} - {f_i}^{(3)}\right|}{b_i - a_i}\right)^{\lambda}\right)^{\frac{1}{\lambda}}$$

$$\leq \left(\sum_{i=1}^{n} \omega_i \left(\frac{\left|{g_i}^{(1)} - {f_i}^{(1)}\right| + \left|{f_i}^{*(1)} - {g_i}^{(1)}\right| + 2\left(\left|{g_i}^{(2)} - {f_i}^{(2)}\right| + \left|{f_i}^{*(2)} - {g_i}^{(2)}\right|\right) + \left|{g_i}^{(3)} - {f_i}^{(3)}\right| + \left|{f_i}^{*(3)} - {g_i}^{(3)}\right|}{b_i - a_i}\right)^{\lambda}\right)^{\frac{1}{\lambda}} \quad (8)$$

According to the Minkowski Inequality [9] we know that:

$$\left(\sum_{i=1}^{n} \omega_i |p_i + q_i|^\lambda\right)^{\frac{1}{\lambda}} \leq \left(\sum_{i=1}^{n} \omega_i |p_i|^\lambda\right)^{\frac{1}{\lambda}} + \left(\sum_{i=1}^{n} \omega_i |q_i|^\lambda\right)^{\frac{1}{\lambda}} \quad (9)$$

Thus, if we consider $p_i = \frac{\left|{g_i}^{(1)} - {f_i}^{(1)}\right| + 2\left|{g_i}^{(2)} - {f_i}^{(2)}\right| + \left|{g_i}^{(3)} - {f_i}^{(3)}\right|}{b_i - a_i}$ and also, we consider $q_i = \frac{\left|{f_i}^{*(1)} - {g_i}^{(1)}\right| + 2\left|{f_i}^{*(2)} - {g_i}^{(2)}\right| + \left|{f_i}^{*(3)} - {g_i}^{(3)}\right|}{b_i - a_i}$, we can rewrite inequality (8) as follows:

$$\left(\sum_{i=1}^{n} \omega_i \left|\frac{\left|{g_i}^{(1)} - {f_i}^{(1)}\right| + \left|{f_i}^{*(1)} - {g_i}^{(1)}\right| + 2\left(\left|{g_i}^{(2)} - {f_i}^{(2)}\right| + \left|{f_i}^{*(2)} - {g_i}^{(2)}\right|\right) + \left|{g_i}^{(3)} - {f_i}^{(3)}\right| + \left|{f_i}^{*(3)} - {g_i}^{(3)}\right|}{b_i - a_i}\right|^{\lambda}\right)^{\frac{1}{\lambda}}$$

$$\leq \left(\sum_{i=1}^{n} \omega_i \left|\frac{\left|{g_i}^{(1)} - {f_i}^{(1)}\right| + 2\left|{g_i}^{(2)} - {f_i}^{(2)}\right| + \left|{g_i}^{(3)} - {f_i}^{(3)}\right|}{b_i - a_i}\right|^{\lambda}\right)^{\frac{1}{\lambda}}$$

$$+ \left(\sum_{i=1}^{n} \omega_i \left|\frac{\left|{f_i}^{*(1)} - {g_i}^{(1)}\right| + 2\left|{f_i}^{*(2)} - {g_i}^{(2)}\right| + \left|{f_i}^{*(3)} - {g_i}^{(3)}\right|}{b_i - a_i}\right|^{\lambda}\right)^{\frac{1}{\lambda}} \quad (10)$$

Then, by joining (8) and (10) we have:

$$\left(\sum_{i=1}^{n} \omega_i \left(\frac{\left|f_i^{*(1)} - f_i^{(1)}\right| + 2\left|f_i^{*(2)} - f_i^{(2)}\right| + \left|f_i^{*(3)} - f_i^{(3)}\right|}{b_i - a_i}\right)^{\lambda}\right)^{\frac{1}{\lambda}}$$

$$\leq \left(\sum_{i=1}^{n} \omega_i \left|\frac{\left|g_i^{(1)} - f_i^{(1)}\right| + 2\left|g_i^{(2)} - f_i^{(2)}\right| + \left|g_i^{(3)} - f_i^{(3)}\right|}{b_i - a_i}\right|^{\lambda}\right)^{\frac{1}{\lambda}}$$

$$+ \left(\sum_{i=1}^{n} \omega_i \left|\frac{\left|f_i^{*(1)} - g_i^{(1)}\right| + 2\left|f_i^{*(2)} - g_i^{(2)}\right| + \left|f_i^{*(3)} - g_i^{(3)}\right|}{b_i - a_i}\right|^{\lambda}\right)^{\frac{1}{\lambda}} \quad (11)$$

Since $b_i > a_i$, the absolute values can be omitted. Then, introducing $\frac{M}{4}$, we have:

$$\frac{M}{4}\left(\sum_{i=1}^{n} \omega_i \left(\frac{\left|f_i^{*(1)} - f_i^{(1)}\right| + 2\left|f_i^{*(2)} - f_i^{(2)}\right| + \left|f_i^{*(3)} - f_i^{(3)}\right|}{b_i - a_i}\right)^{\lambda}\right)^{\frac{1}{\lambda}}$$

$$\leq \frac{M}{4}\left(\sum_{i=1}^{n} \omega_i \left(\frac{\left|g_i^{(1)} - f_i^{(1)}\right| + 2\left|g_i^{(2)} - f_i^{(2)}\right| + \left|g_i^{(3)} - f_i^{(3)}\right|}{b_i - a_i}\right)^{\lambda}\right)^{\frac{1}{\lambda}}$$

$$+ \frac{M}{4}\left(\sum_{i=1}^{n} \omega_i \left(\frac{\left|f_i^{*(1)} - g_i^{(1)}\right| + 2\left|f_i^{*(2)} - g_i^{(2)}\right| + \left|f_i^{*(3)} - g_i^{(3)}\right|}{b_i - a_i}\right)^{\lambda}\right)^{\frac{1}{\lambda}} \quad (12)$$

Consequently:

$$d_{TR}\left(\tilde{f}, \tilde{f}^{*}\right) \leq d_{TR}\left(\tilde{f}, \tilde{g}\right) + d_{TR}\left(\tilde{g}, \tilde{f}^{*}\right)$$

And $d_{TR}$ is a *distance function* according to ***Definition 4*** ■

**Proposition 2 (Bounded).** Let $(\mathcal{F}, d_{TR})$ be a non-empty metric space where $\mathcal{F}$ is defined as above. Then $d_{TR}$ normalizes the distance between two tuples of Triangular Fuzzy Numbers (TFNs) within the range [0, M].

*Proof*. We need to show that $d_{TR}$ falls within the range [0, M] for any two *n*-tuples of TFNs in $\mathcal{F}$. First, let´s consider the term inside the sum:

$$\frac{\left|f_i^{*(1)} - f_i^{(1)}\right| + 2\left|f_i^{*(2)} - f_i^{(2)}\right| + \left|f_i^{*(3)} - f_i^{(3)}\right|}{b_i - a_i}$$

Since $f_i^{(k)}$ and $f_i^{*(k)}$ are both in the interval $[a_i, b_i]$, their difference is bounded by $\left|f_i^{*(k)} - f_i^{(k)}\right| \leq b_i - a_i$. Therefore, the numerator is at most $(b_i - a_i) + 2(b_i - a_i) + (b_i - a_i)$= 4$(b_i - a_i)$, leading to:

$$\frac{\left|f_i^{*(1)} - f_i^{(1)}\right| + 2\left|f_i^{*(2)} - f_i^{(2)}\right| + \left|f_i^{*(3)} - f_i^{(3)}\right|}{b_i - a_i} \leq 4$$

Thus, each term inside the sum is at most 4. Given that $\omega_i \in [0; 1]$ and $\sum_{i=1}^{n} \omega_i = 1$, we have:

$$\sum_{i=1}^{n}\omega_i\left(\frac{\left|f_i^{*(1)}-f_i^{(1)}\right|+2\left|f_i^{*(2)}-f_i^{(2)}\right|+\left|f_i^{*(3)}-f_i^{(3)}\right|}{b_i-a_i}\right)^{\lambda}\leq 4^{\lambda}$$

Taking the $\frac{1}{\lambda}$ power:

$$\left(\sum_{i=1}^{n}\omega_i\left(\frac{\left|f_i^{*(1)}-f_i^{(1)}\right|+2\left|f_i^{*(2)}-f_i^{(2)}\right|+\left|f_i^{*(3)}-f_i^{(3)}\right|}{b_i-a_i}\right)^{\lambda}\right)^{\frac{1}{\lambda}}\leq 4$$

Multiplying by $\frac{M}{4}$:

$$\frac{M}{4}\left(\sum_{i=1}^{n}\omega_i\left(\frac{\left|f_i^{*(1)}-f_i^{(1)}\right|+2\left|f_i^{*(2)}-f_i^{(2)}\right|+\left|f_i^{*(3)}-f_i^{(3)}\right|}{b_i-a_i}\right)^{\lambda}\right)^{\frac{1}{\lambda}}\leq M$$

Then, $d_{TR}\leq M$. For the lower bound, since all terms are non-negative, we have that $d_{TR}\geq 0$. Thus, $d_{TR}$ is bounded within [0, M]■.

**Proposition 3 (Scale Invariance).** Let $(\mathcal{F}, d_{TR})$ be a non-empty metric space where $\mathcal{F}$ is defined as above. Then, the Triangular Fuzzy Rescaling Distance $d_{TR}$ is scale-invariant.

*Proof.* To prove scale invariance, we need to show that the distance measure $d_{TR}$ remains unchanged when both the reference and target Triangular Fuzzy Numbers (TFNs) are scaled by a positive constant $k$, and the intervals are scaled accordingly:

- Scaled TFNs: $\tilde{f}_i^{*\prime}=\left(kf_i^{*(1)},kf_i^{*(2)},kf_i^{*(3)}\right)$ and $\tilde{f}_i^{\prime}=\left(kf_i^{(1)},kf_i^{(2)},kf_i^{(3)}\right)$
- Scaled intervals: $[ka_i,kb_i]$, so $b_i-a_i$ becomes $k(b_i-a_i)$

Then $d'_{TR}$ is given by:

$$d'_{TR}=\frac{M}{4}\left(\sum_{i=1}^{n}\omega_i\left(\frac{\left|kf_i^{*(1)}-kf_i^{(1)}\right|+2\left|kf_i^{*(2)}-kf_i^{(2)}\right|+\left|kf_i^{*(3)}-kf_i^{(3)}\right|}{k(b_i-a_i)}\right)^{\lambda}\right)^{\frac{1}{\lambda}}$$

$$=\frac{M}{4}\left(\sum_{i=1}^{n}\omega_i\left(\frac{k\left(\left|f_i^{*(1)}-f_i^{(1)}\right|+2\left|f_i^{*(2)}-f_i^{(2)}\right|+\left|f_i^{*(3)}-f_i^{(3)}\right|\right)}{k(b_i-a_i)}\right)^{\lambda}\right)^{\frac{1}{\lambda}}$$

$$=\frac{M}{4}\left(\sum_{i=1}^{n}\omega_i\left(\frac{\left|f_i^{*(1)}-f_i^{(1)}\right|+2\left|f_i^{*(2)}-f_i^{(2)}\right|+\left|f_i^{*(3)}-f_i^{(3)}\right|}{b_i-a_i}\right)^{\lambda}\right)^{\frac{1}{\lambda}}$$

Hence, $d_{TR}$ is scale-invariant ■.

**Proposition 4 (Origin Invariance).** Let $(\mathcal{F}, d_{TR})$ be a non-empty metric space where $\mathcal{F}$ is defined as above. Then, the Triangular Fuzzy Rescaling Distance $d_{TR}$ is origin-invariant.

*Proof.* To prove that $d_{TR}$ is origin-invariant, we need to show that the distance between two Triangular Fuzzy Numbers (TFNs) remains unchanged when both are shifted by the same constant $\phi$. Consider two $n$-tuples of TFNs $\tilde{f}_i=\left(f_i^{(1)},f_i^{(2)},f_i^{(3)}\right)$

and $\tilde{f}_i^* = \left(f_i^{*(1)}, f_i^{*(2)}, f_i^{*(3)}\right)$ defined over the interval $[a_i, b_i]$. If we shift the TFNs and their intervals by a positive constant $\phi$:

- Scaled TFNs: $\tilde{f}_i^* + \phi = \left(f^{*(1)} + \phi, f^{*(2)} + \phi, f^{*(3)} + \phi\right)$ and $\tilde{f}_i + \phi = \left(f^{(1)} + \phi, f^{(2)} + \phi, f^{(3)} + \phi\right)$
- Scaled interval length: $(b_i + \phi) - (a_i + \phi)$

Then, we have:

$$d'_{TR} = \frac{M}{4}\left(\sum_{i=1}^{n} \omega_i \left(\frac{\left|\left(f_i^{*(1)} + \phi\right) - \left(f_i^{(1)} + \phi\right)\right| + 2\left|\left(f_i^{*(2)} + \phi\right) - \left(f_i^{(2)} + \phi\right)\right| + \left|\left(f_i^{*(3)} + \phi\right) - \left(f_i^{(3)} + \phi\right)\right|}{(b_i + \phi) - (a_i + \phi)}\right)^{\lambda}\right)^{\frac{1}{\lambda}}$$

$$d'_{TR} = \frac{M}{4}\left(\sum_{i=1}^{n} \omega_i \left(\frac{\left|f_i^{*(1)} - f_i^{(1)}\right| + 2\left|f_i^{*(2)} - f_i^{(2)}\right| + \left|f_i^{*(3)} - f_i^{(3)}\right|}{b_i - a_i}\right)^{\lambda}\right)^{\frac{1}{\lambda}} = d_{TR}$$

Therefore, $d_{TR}$ is origin-invariant ■.

**Proposition 5 (Normalization).** Let $(\mathcal{F}, d_{TR})$ be a non-empty metric space where $\mathcal{F}$ is defined as above. Then, the Triangular Fuzzy Rescaling Distance $d_{TR}$ normalizes the characteristic values $\left(f_i^{(1)}, f_i^{(2)}, f_i^{(3)}\right)$ and $\left(f_i^{*(1)}, f_i^{*(2)}, f_i^{*(3)}\right)$ of each pair of $n$-tuples of triangular fuzzy numbers $\tilde{f}_i^*, \tilde{f}_i$ by Linear Rescaling (LRE).

*Proof.* Upon transforming the $d_{TR}$ we have:

$$d_{TR} = \frac{M}{4}\left(\sum_{i=1}^{n} \omega_i \left(\frac{\left|f_i^{*(1)} - f_i^{(1)}\right| + 2\left|f_i^{*(2)} - f_i^{(2)}\right| + \left|f_i^{*(3)} - f_i^{(3)}\right|}{b_i - a_i}\right)^{\lambda}\right)^{\frac{1}{\lambda}}$$

$$= \frac{M}{4}\left(\sum_{i=1}^{n} \omega_i \left(\left|\frac{f_i^{*(1)} - f_i^{(1)} + b_i - b_i}{b_i - a_i}\right| + 2\left|\frac{f_i^{*(2)} - f_i^{(2)} + b_i - b_i}{b_i - a_i}\right| + \left|\frac{f_i^{*(3)} - f_i^{(3)} + b_i - b_i}{b_i - a_i}\right|\right)^{\lambda}\right)^{\frac{1}{\lambda}}$$

$$= \frac{M}{4}\left(\sum_{i=1}^{n} \omega_i \left(\left|\frac{b_i - f_i^{(1)}}{b_i - a_i}\right| - \left|\frac{b_i - f_i^{*(1)}}{b_i - a_i}\right| + 2\left|\frac{b_i - f_i^{(2)}}{b_i - a_i}\right| - 2\left|\frac{b_i - f_i^{*(2)}}{b_i - a_i}\right| + \left|\frac{b_i - f_i^{(3)}}{b_i - a_i}\right| - \left|\frac{b_i - f_i^{*(3)}}{b_i - a_i}\right|\right)^{\lambda}\right)^{\frac{1}{\lambda}}$$

$$= \frac{1}{4}\left(\sum_{i=1}^{n} \omega_i \left(M\left|\frac{b_i - f_i^{(1)}}{b_i - a_i}\right| - M\left|\frac{b_i - f_i^{*(1)}}{b_i - a_i}\right| + 2M\left|\frac{b_i - f_i^{(2)}}{b_i - a_i}\right| - 2M\left|\frac{b_i - f_i^{*(2)}}{b_i - a_i}\right| + M\left|\frac{b_i - f_i^{(3)}}{b_i - a_i}\right| - M\left|\frac{b_i - f_i^{*(3)}}{b_i - a_i}\right|\right)^{\lambda}\right)^{\frac{1}{\lambda}}$$

According to **Proposition 2**: $0 \le d_{TR} \le M$, then: $0 = min(d_{TR}) = m$ and $max(d_{TR}) = M$, thus:

$$d_{TR} = \frac{1}{4}\left(\sum_{i=1}^{n} \omega_i \left((M - m)\left|\frac{b_i - f_i^{(1)}}{b_i - a_i}\right| - (M - m)\left|\frac{b_i - f_i^{*(1)}}{b_i - a_i}\right| + 2\left((M - m)\left|\frac{b_i - f_i^{(2)}}{b_i - a_i}\right| - (M - m)\left|\frac{b_i - f_i^{*(2)}}{b_i - a_i}\right|\right) + (M - m)\left|\frac{b_i - f_i^{(3)}}{b_i - a_i}\right| - (M - m)\left|\frac{b_i - f_i^{*(3)}}{b_i - a_i}\right|\right)^{\lambda}\right)^{\frac{1}{\lambda}}$$

Adding a $M - M = 0$:

$$d_{TR} = \frac{1}{4}\left(\sum_{i=1}^{n} \omega_i \left(M + (M-m)\left|\frac{b_i - f_i^{(1)}}{b_i - a_i}\right| - M - (M-m)\left|\frac{b_i - f_i^{*(1)}}{b_i - a_i}\right| + 2\left(M + (M-m)\left|\frac{b_i - f_i^{(2)}}{b_i - a_i}\right| - M - (M-m)\left|\frac{b_i - f_i^{*(2)}}{b_i - a_i}\right|\right) + M + (M-m)\left|\frac{b_i - f_i^{(3)}}{b_i - a_i}\right| - M - (M-m)\left|\frac{b_i - f_i^{*(3)}}{b_i - a_i}\right|\right)^{\lambda}\right)^{\frac{1}{\lambda}}$$

$$= \frac{1}{4}\left(\sum_{i=1}^{n} \omega_i \left[M - (M-m)\left|\frac{b_i - f_i^{*(1)}}{b_i - a_i}\right| - \left(M - (M-m)\left|\frac{b_i - f_i^{(1)}}{b_i - a_i}\right|\right) + 2\left(\left[M - (M-m)\left|\frac{b_i - f_i^{*(2)}}{b_i - a_i}\right| - \left(M - (M-m)\left|\frac{b_i - f_i^{(2)}}{b_i - a_i}\right|\right)\right]\right) + M - (M-m)\left|\frac{b_i - f_i^{*(3)}}{b_i - a_i}\right| - \left(M - (M-m)\left|\frac{b_i - f_i^{(3)}}{b_i - a_i}\right|\right)\right]^{\lambda}\right)^{\frac{1}{\lambda}}$$

$$d_{TR} = \frac{1}{4}\left(\sum_{i=1}^{n} \omega_i \left([LRE(f_i^{*(1)}) - LRE(f_i^{(1)})] + 2[LRE(f_i^{*(2)}) - LRE(f_i^{(2)})] + [LRE(f_i^{*(3)}) - LRE(f_i^{(3)})]\right)^{\lambda}\right)^{\frac{1}{\lambda}} \blacksquare$$

# 4 Numerical Example

Let's consider an example with $n$=2 (two-dimensional tuples of TFNs) and specific values for the intervals and TFNs: $IN = \{[1.00, 3.00], [-10.00, 0.00]\}$; $\widetilde{f_1} = (1.00, 1.50, 2.00)$; $\widetilde{f_1}^* = (1.50, 2.00, 2.50)$; $\widetilde{f_2} = (-10.00, -9.50, -7.00)$; $\widetilde{f_2}^* = (-2.00, -1.50, -1.00)$; $\omega_1 = 0.50$; $\omega_2 = 0.50$; $\lambda = 2; M = 100$. To calculate the scaled distance $d_{TR}$ between them, we must follow these steps:

***Step 1****. Calculate the difference for each dimension (i.e., each pair of TFN):*

For $\widetilde{f_1}$ and $\widetilde{f_1}^*$:

$$\frac{|f_1^{*(1)} - f_1^{(1)}| + 2|f_1^{*(2)} - f_1^{(2)}| + |f_1^{*(3)} - f_1^{(3)}|}{b_1 - a_1} = \frac{|1.5 - 1.0| + 2|2.0 - 1.5| + |2.5 - 2.0|}{3.0 - 1.0} = \mathbf{1.00}$$

For $\widetilde{f_2}$ and $\widetilde{f_2}^*$:

$$\frac{|f_1^{*(1)} - f_1^{(1)}| + 2|f_1^{*(2)} - f_1^{(2)}| + |f_1^{*(3)} - f_1^{(3)}|}{b_1 - a_1} = \frac{|-10 + 2.0| + 2|-9.5 + 1.5| + |-7.0 + 1.0|}{0.0 - (-10.0)} = \mathbf{3.00}$$

***Step 2****. Apply the weighting vector:*

$$\sum_{i=1}^{2} \omega_i \left(\frac{|f_i^{*(1)} - f_i^{(1)}| + 2|f_i^{*(2)} - f_i^{(2)}| + |f_i^{*(3)} - f_i^{(3)}|}{b_i - a_i}\right)^{\lambda} = 0.5 \cdot 1.0^2 + 0.5 \cdot 3.0^2 = \mathbf{5.00}$$

***Step 3****. Calculate the final distance:* $d_{TR}\left((\widetilde{f_1}, \widetilde{f_2}), (\widetilde{f_1}^*, \widetilde{f_2}^*)\right) = \frac{100}{4}(5)^{\frac{1}{2}} = \mathbf{55.90}$

If we repeat the same steps considering that $\lambda$=1, we obtain $d_{TR}$= 50.00

## 5 An Application to Fuzzy Multi-Criteria Decision Making

Assume a regional initiative aiming to assess cities based on their sustainability performance. We will evaluate five hypothetical cities (C1, C2, C3, C4, and C5) based on five strategic sustainability indicators (S1, S2, S3, S4, S5):

**S1:** CO2 Emissions per capita (tonnes/year). Range: [0, 20] tonnes/year. Scale: Lower is better.

**S2**: Access to Green Space per capita (square meters/capita). Range: [0, 500] sq m/capita. Scale: Higher is better.

**S3**: Public Transportation Usage Rate (% of commuters). Range: [0, 100]. Scale: Higher is better.

**S4**: Income Inequality (Gini Index). Range: [0, 1]. Scale: Lower is better.

**S5**: Waste Recycling Rate (% of total waste). Range: [0, 100]. Scale: Higher is better.

In this problem, the goal is to determine how far each city is from ideal levels of sustainability. The fuzzy value for each city, expressed in TFNs, is shown in **Table 1,** together with the ideal TFNs.

**Table 1.** Data for Cities

| | S1 | S2 | S3 | S4 | S5 |
|---|---|---|---|---|---|
| **C1** | (14, 15, 20) | (0, 50, 70) | (0, 5, 10) | (0.87, 0.93, 1.00) | (0, 5, 10) |
| **C2** | (20, 20, 20) | (480, 490, 500) | (95, 97, 100) | (0.00, 0.02, 0.03) | (90, 95, 99) |
| **C3** | (0, 1, 2) | (480, 490, 500) | (95, 97, 100) | (0.00, 0.02, 0.03) | (90, 95, 99) |
| **C4** | (9, 10, 11) | (100, 150, 200) | (30, 40, 50) | (0.50, 0.55, 0.60) | (20, 30, 40) |
| **C5** | (4, 5, 6) | (200, 220, 240) | (55, 60, 65) | (0.30, 0.32, 0.34) | (40, 45, 50) |
| **Ideal** | (0, 0, 0) | (500, 500, 500) | (100, 100, 100) | (0.00, 0.00, 0.00) | (100, 100, 100) |

We will analyze the results obtained using the proposed $d_{TR}$ (with M=100, across varying λ parameters. We aggregate this information in order to make a decision (**Table 2**). Equal weights will be maintained for all strategic indicators, for simplicity: **Ω** = (0.2, 0.2, 0.2, 0.2, 0.2).

**Table 2.** Aggregated results

| | λ=1 | λ=2 | λ=3 | λ=500 |
|---|---|---|---|---|
| **C1** | 90.95 | 91.12 | 91.29 | 94.83 |
| **C2** | 22.35 | 44.82 | 58.48 | 99.68 |
| **C3** | 3.35 | 3.67 | 3.93 | 0.00 |
| **C4** | 61.00 | 61.52 | 62.03 | 69.87 |
| **C5** | 41.60 | 43.38 | 44.96 | 55.82 |
| **Ranking** | **C3≻C2≻C5 ≻C4≻C1** | **C3≻C5≻C2 ≻C4≻C1** | **C3≻C5≻C2 ≻C4≻C1** | **C3≻C5≻C4 ≻C1≻C2** |

Note that ≻ means preferred to.

The $d_{TR}$ metric yields distance values within the [0, $M$] interval, irrespective of whether small or large values of λ are used (λ=1, 2, 3, and a significantly higher λ=500). This inherent normalization feature of $d_{TR}$ allows for a direct interpretation of its values as a proportional deviation from the ideal. This characteristic reinforces the utility of $d_{TR}$ for raw data (**Proposition 5**) and could be of great practical use, for example, in the formulation of synthetic indicators based on distance measurements. For instance, considering the case of λ=1 as a linear baseline, the $d_{TR}$ value for city C1 is 90.95, suggesting that C1 is 90.95% distant from the combined ideal sustainability levels across the five strategic indicators, implying it has achieved roughly 9.05% of the ideal sustainability in this context. Conversely, C3, with a $d_{TR}$ of only 3.35 for λ=1, appears remarkably close to the ideal, being only about 3.35% away from perfect sustainability.

Interestingly, the ranking of cities shifts depending on the λ value (**Table 2**). At higher values of $\lambda$, such as λ=500, the $d_{TR}$ metric becomes more sensitive to the largest discrepancies. While lower values of $\lambda$ (like λ=1,2) offer a more balanced aggregation of deviations across all indicators, higher values of $\lambda$ may be particularly relevant when decision-makers prioritize the identification and mitigation of critical weaknesses (e.g., when a city performs well on most indicators, but shows a serious deficiency in one crucial area). For example, comparing C2 and C3 is key to understanding this effect: they have identical data except for S1 (CO2), where C2 performs worse (**Table 1**). When incrementing λ, the $d_{TR}$ values for C2 increase drastically from 22.35 to 99.68 (C2's relatively poor S1 score becomes heavily penalized at λ=500 because, in the context of sustainability, equilibrium across factors is vital). Calculating the $d_{TR}$ metric with several λ values helps to detect possible imbalances even if the weights between indicators vary.

## 6 Conclusion

This paper introduced the Triangular Fuzzy Rescaling Distance ($d_{TR}$), a novel metric designed for comparing n-tuples of Triangular Fuzzy Numbers (TFNs). We formally demonstrated that $d_{TR}$ satisfies the properties of a metric and, importantly, is bounded, scale-invariant, and origin-invariant. Its key feature is the direct integration of Linear Rescaling (LRE) into the distance calculation. This inherent normalization addresses the challenge of comparing fuzzy data across heterogeneous scales and units without requiring separate preprocessing steps, while maintaining linear O(n) computational complexity.

The properties of $d_{TR}$ suggest its utility in applications involving imprecise or vague information, particularly in constructing synthetic indicators. By quantifying the distance between observed and target states represented as TFNs, $d_{TR}$ can serve as an aggregation mechanism for multi-dimensional concepts (e.g., sustainability, quality of life, etc.). Its bounded nature aids interpretability, and the inclusion of a weighting vector allows for prioritizing dimensions. Furthermore, it could also be of interest in distance-based decision methods like TOPSIS [10] or clustering and classification [6].

While promising, this study has limitations. The current formulation of $d_{TR}$ is specifically defined for TFNs, which might restrict its direct applicability to scenarios where other types of fuzzy representations, such as trapezoidal or Gaussian fuzzy numbers, are more appropriate or prevalent. Furthermore, interpreting the resulting distance values requires careful consideration within the problem's domain context. Future research should focus on empirically validating $d_{TR}$ in diverse real-world scenarios, comparing its performance against existing distance measures. Exploring application-specific parameter tuning strategies, extending the approach to other types of fuzzy numbers, and developing an open-source implementation are also valuable directions to facilitate broader adoption.

**Acknowledgments.** This research received support from the Agència de Gestió d'Ajuts Universitaris i de Recerca (AGAUR) and was co-funded by the European Union, under Grant 2023 FI-1 00622, and takes part of the Grant PID2021-122575NB-I00 funded by MCIN/AEI/10.13039/501100011033/ by "ERDF A way of making Europe". This work was also partially funded by Universitat Rovira i Virgili (URV) [project number 2023PFR-URV-114], and by Generalitat de Catalunya (AGAUR) to ITAKA consolidated research group [2021-SGR-00114]. The work was also supported by the Spanish Network ELIGE-IA (RED2022-134302-T).